\documentclass{article}
\usepackage{arxiv}
\usepackage[utf8]{inputenc}

\usepackage{amssymb}
\usepackage{amsmath}

\usepackage[colorlinks=true,linkcolor=blue,hidelinks,breaklinks]{hyperref}  
    
\usepackage{algorithm}
\usepackage{algpseudocode}
\usepackage{caption}
\usepackage{subcaption}
\usepackage{multicol}
\usepackage{multirow}
\usepackage{graphicx}
\usepackage{orcidlink}
\usepackage{pdfpages}
\usepackage{threeparttable}
\usepackage{booktabs}

\usepackage{acronym}
\acrodef{ai}[AI]{artificial intelligence}
\acrodef{asic}[ASIC]{application-specific integrated circuit}
\acrodef{bptt}[BPTT]{backpropagation through time}
\acrodef{bpttc}[BPTT-C]{BPTT with checkpoints}
\acrodef{fecap}[FeCap]{ferroelectric capacitor}
\acrodef{felif}[FeLIF]{ferroelectric leaky integrate-and-fire}
\acrodef{lif}[LIF]{leaky integrate-and-fire}
\acrodef{li}[LI]{leaky integrator}
\acrodef{if}[IF]{integrate-and-fire}
\acrodef{C2C}[C2C]{Cycle-to-Cycle}
\acrodef{VCM}[VCM]{valence change mechanism}
\acrodef{OE}[OE]{ohmic electrode}
\acrodef{AE}[AE]{active electrode}
\acrodef{TE}[TE]{top electrode}
\acrodef{BE}[BE]{bottom electrode}
\acrodef{HRS}[HRS]{high resistive state}
\acrodef{LRS}[LRS]{low resistive states}
\acrodef{DPI}[DPI]{differential pair integrator}
\acrodef{ste}[STE]{straight-through estimators}
\acrodef{snn}[SNN]{Spiking Neural Network}
\acrodef{hpo}[HPO]{Hyperparameter Optimization}
\acrodef{qat}[QAT]{Quantisation Aware Training}
\acrodef{JART}[JART]{Jülich Aachen Resistive Switching Tool}
\acrodef{RRAM}[RRAM]{Resistive random-access memory}
\acrodef{MC}[MC]{Monte Carlo}
\acrodef{cmos}[CMOS]{complementary metal-oxide semiconductor}
\acrodef{bruno}[BRUNO]{Backpropagation Running Undersampled for Novel device Optimization}
\acrodef{ml}[ML]{Machine Learning}
\acrodef{gpu}[GPU]{graphic processing unit}
\acrodef{pcm}[PCM]{phase-change memory}
\acrodef{sram}[SRAM]{static random-access memory}
\acrodef{jsb}[JSB]{Johann Sebastian Bach}
\acrodef{aihwkit}[AIHWKIT]{IBM Analog Hardware Acceleration Kit}
\acrodef{spice}[SPICE]{Simulation Program with Integrated Circuit Emphasis}

\usepackage{authblk}

\author[1]{\orcidlink{0000-0003-0993-5593}Luca Fehlings}
\author[1]{\orcidlink{0009-0004-4089-4528}Bojian Zhang}
\author[1]{\orcidlink{0009-0002-6006-608X}Paolo Gibertini}
\author[1]{Martin A. Nicholson}
\author[1]{\orcidlink{0000-0003-0479-6897}Erika Covi}
\author[1]{\orcidlink{0000-0001-5042-9399}Fernando M. Quintana}

\affil[1]{Zernike Institute for Advanced Materials \& Groningen Cognitive Systems and Materials Center (CogniGron)
	University of Groningen, 9747 AG Groningen, Netherlands}

\renewcommand{\headeright}{}
\renewcommand{\undertitle}{}
\renewcommand{\shorttitle}{Bruno}

\hypersetup{
pdftitle={Bruno: Backpropagation Running Undersampled for Novel device Optimization},
pdfauthor={Luca Fehlings, Bojian Zhang, Paolo Gibertini, Martin A. Nicholson, Erika Covi, Fernando M. Quintana},
pdfkeywords={Learning, FeCAP, neuromorphic, SNN, RRAM},
}

\usepackage{pdfpages}

\title{Bruno: Backpropagation Running Undersampled for Novel device Optimization}
    
\begin{document}

\maketitle

\begin{abstract}
Recent efforts to improve the efficiency of neuromorphic and machine learning systems have centred on developing of specialised hardware for neural networks. These systems typically feature architectures that go beyond the von Neumann model employed in general-purpose hardware such as GPUs, offering potential efficiency and performance gains. However, neural networks developed for specialised hardware must consider its specific characteristics. This requires novel training algorithms and accurate hardware models, since they cannot be abstracted as a general-purpose computing platform. In this work, we present a bottom-up approach to training neural networks for hardware-based spiking neurons and synapses, built using ferroelectric capacitors (FeCAPs) and resistive random-access memories (RRAMs), respectively. Unlike the common approach of designing hardware to fit abstract neuron or synapse models, we start with compact models of the physical device to model the computational primitives. Based on these models, we have developed a training algorithm (BRUNO) that can reliably train the networks, even when applying hardware limitations, such as stochasticity or low bit precision. We analyse and compare BRUNO with Backpropagation Through Time. We test it on different spatio-temporal datasets. First on a music prediction dataset, where a network composed of ferroelectric leaky integrate-and-fire (FeLIF) neurons is used to predict at each time step the next musical note that should be played. The second dataset consists on the classification of the Braille letters using a network composed of quantised RRAM synapses and FeLIF neurons. The performance of this network is then compared with that of networks composed of LIF neurons. Experimental results show the potential advantages of using BRUNO by reducing the time and memory required to detect spatio-temporal patterns with quantised synapses.
\end{abstract}



\section{Introduction} 
    \label{sec:intro}
    The rapid growth of \ac{ai} demands efficient hardware solutions that enable energy-efficient computing without sacrificing performance. However, current \ac{ai} hardware faces fundamental trade-offs between power consumption, computational efficiency, and accuracy. Furthermore, the storage and routing of synaptic weights remains one of the major challenges for efficient hardware platforms for \acp{snn} \cite{yan2020low}. While advantageous, quantisation of synaptic weights can introduce noise and nonlinearity that degrade the network's accuracy performance in inference. Neurons using innovative devices promise better noise rejection \cite{gibertini2024} and, as we will show, allow strict weight quantisation. Moreover, memory devices have shown potential in their application in \ac{ml} and neuromorphic engineering for in-memory computing \cite{ielmini2018memory} due to their non-volatility, compact size, and low latency \cite{10052010}. Thanks to these properties, devices such as \ac{RRAM}, \ac{pcm} or ferroelectric devices offer situational advantages over traditional memories such as \ac{sram}. Their non-volatility makes it possible to store the parameters of \ac{ml} models without a constant power supply. This characteristic makes them ideal for representing static and plastic synapses of an artificial neural network.
    
    However, the use of these memory devices within an \acp{asic} requires accurate physical or behavioural models. These models typically represent the network synapses, where their behavioural models are incorporated into simulation frameworks and training algorithms \cite{gokmen2020algorithm,demirag2023overcoming,10548824,demirag2021online,10019455,10.3389/frai.2021.692065} to account for device variability in the updating process. Recent research has also incorporated physical models into machine learning frameworks such as \ac{aihwkit} \cite{10052010,aihwkit}. However, these devices can also be used as computational primitives in the neuron to enhance its behaviour by including multiple time constants \cite{gibertini2024}. When used as computational primitives, these models have to be computationally efficient and well-posed to allow efficient temporal training methods, such as \ac{bptt} \cite{8891809}. Accurate models and efficient training are particularly important when the inherent physical dynamics of the underlying hardware are to be exploited. Algorithms and hardware need to be co-developed from memory devices and circuit technology to the whole neuromorphic system.
    
    One of the challenges of using memory devices is that they are inherently stateful and exhibit non-linear behaviour over a wide range of timescales that need to be reliably modelled. Given that typical application timescales are on the order of 100 milliseconds to seconds \cite{9311226,see2020st,8747378}, integrating these models, whose simulation timescale is on the order of microseconds, with gradient-based training algorithms, such as \ac{bptt}, would lead to a significant increase in training time and memory requirements. This is because the computational graph created for automatic differentiation contains the states at every microsecond. An alternative approach would be to increase the time step, but this would lead to model instability and inaccurate results. Therefore, we propose an innovative training method that reduces the training time by around an order of magnitude and reduces the memory consumption during training over a large number of time steps. This is achieved by using two different time resolutions for the forward and the backward passes. We call this method \ac{bruno}.
    This training method has several advantages, namely reducing time and memory consumption compared to \ac{bptt}, allowing us to train neural networks with high temporal dynamics, and compare the behaviour of networks with these new models with standard neurons such as \ac{lif} \cite{gerstner2002spiking}. To validate \ac{bruno}, a network composed of \ac{felif} neurons and synapses based on \ac{RRAM} devices is trained and compared with \ac{lif} neurons. In addition, thanks to the \ac{bruno} algorithm, we can showcase how hardware-based models, like the \ac{felif}, could outperform or match recurrent neural networks when the synaptic weights are of low bit precision.

\section{Materials and Methods} 
    \label{sec:methods}
    
    \subsection{Ferroelectric LIF neuron}
        \label{sec:methods:felif}
\begin{figure}[t]
            \begin{subfigure}[b]{0.4\textwidth}
                \centering
                \includegraphics[width=\textwidth]{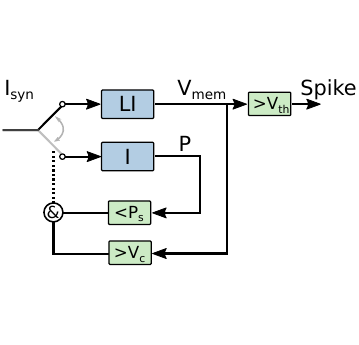}
                \caption{\label{fig:felif_diagram:a}}
            \end{subfigure}
            \begin{subfigure}[b]{0.59\textwidth}
                \centering
                \includegraphics[width=\textwidth]{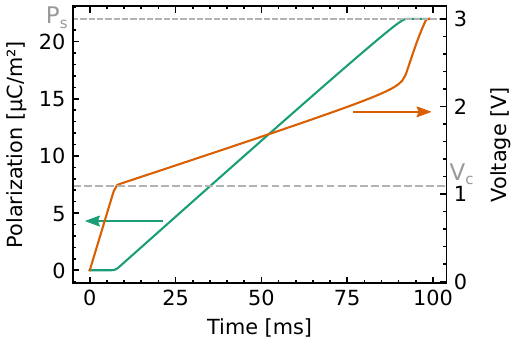}
                \caption{}
            \end{subfigure}
            \caption{(a) Operating principle of the \ac{felif} neuron. When the membrane potential V is charged by the input current I$_{syn}$ via the leaky integrator (LI) and reaches the coercive voltage V$_c$, the current partially flows to the integrator (I). When the integrator state P reaches the saturation value P$_s$, the current flows back to the leaky integrator until it reaches the firing threshold and the neuron emits a spike. (b) Polarization and voltage of the neuron for a constant input current. As described in the operating principle, the input current is integrated into the polarization as long as $V\mathrm{_{mem}}>V\mathrm{_c}$ and $P<P\mathrm{_s}$}
            \label{fig:felif_diagram}
        \end{figure}

        The \ac{felif} neuron \cite{gibertini2024}, which co-integrates \ac{cmos} and \ac{fecap} technology, uses the non-linear charge-voltage relationship of the \ac{fecap} to integrate the inputs to two different state variables, the dielectric and the ferroelectric polarization. The dielectric polarization is linear with the voltage across the capacitor and would result in the behaviour of a conventional \ac{lif} neuron. However, the non-volatile ferroelectric polarization is gated by the membrane potential V$\mathrm{_{mem}}$, the voltage across the capacitor, and only increases when a certain membrane potential V$\mathrm{_c}$ is reached. The ferroelectric polarization then accumulates until the saturation polarization P$\mathrm{_s}$ is reached. The \ac{felif} can thus be described phenomenologically by an \ac{if} neuron gated by a \ac{li} (Figure \ref{fig:felif_diagram:a}), resulting in a neuron capable of tracking significant inputs over long time frames.

        Based on the model extracted from the \ac{fecap}-\ac{cmos} neuron circuit \cite{gibertini2024}, the behaviour can be described as follows.
        The derivative of the membrane potential is obtained by dividing all the currents flowing into the \ac{fecap} by its capacitance and the parasitic capacitance originating from the \ac{cmos} circuit:   
        \begin{equation}
            \frac{dV\mathrm{_{mem}}}{dt} = \frac{I_\mathrm{syn}-I_\mathrm{leak}-I_p}{C_\mathrm{0}+C_\mathrm{par}}
            \label{eq:v_derivative_circuit}
        \end{equation}    
        similar to a \ac{lif} neuron, where V$\mathrm{_{mem}}$ is the membrane potential, $\mathrm{I_{syn}}$ is the synaptic input current, $\mathrm{I_{leak}}$ is the leakage current through the ferroelectric capacitor and $\mathrm{I_p}$ is the displacement current of the ferroelectric polarization. The ferroelectric displacement current results from the ferroelectric polarization process under an electric field and is proportional to the derivative of the polarization:
        \begin{equation}
            \frac{I_p}{A} = \frac{dP}{dt} = \frac{ \textrm{sign}(E_{fe}) \cdot P_\textrm{s}-P}{\tau(E_{fe})}
            \label{eq:dpdt}
        \end{equation}
        where $A$ is the area of the device, $\mathrm{P}$ is the ferroelectric polarization, $E_{fe}$ is the electric field across the ferroelectric layer and $\mathrm{P_s}$ is the saturation polarization. $\mathrm{\tau}$ is the time constant of the polarization and is a function of the electric field:     
        \begin{equation}
            \tau(E) = \tau_0 \cdot \exp\left(\frac{E_\textrm{a}}{|E_{fe}|}\right)^\alpha
            \label{eq:tau}
        \end{equation}
        with activation electric field $\mathrm{E_a}$, elementary time constant $\mathrm{\tau_0}$ and the fitting factor $\mathrm{\alpha}$. The parameters described above used for the simulations are listed in Table \ref{tab:felif_parameters}. 

        \begin{table}[t]
            \caption{Physical parameters of the \ac{felif} neuron model used for simulation.}\label{tab:felif_parameters}
            \centering
            \begin{tabular}{l r}
                \toprule
                Parameter & Value \\
                \midrule
                A & 25\,\textmu m\textsuperscript{2} \\
                C\textsubscript{0} & 0.558\,pF \\
                C\textsubscript{par} & 15\,fF \\
                P\textsubscript{S}& 0.22\,C/m\textsuperscript{2}\\
                E\textsubscript{a}& 1.27\,V/nm\\
                $\tau_0$ & 0.1\,ps\\
                $\alpha$ & 1.3 \\
                \bottomrule
            \end{tabular}
        \end{table}

    \subsection{Resistive switching non-volatile synapses} 
        \label{sec:methods:RRAM}        
        In this work, the weight quantisation is motivated by the use of \ac{RRAM} as n-bit synaptic connections within the network. The \ac{RRAM} switching behaviour is simulated using the \ac{JART} \ac{VCM} physical compact model \cite{bengeljart}. A detailed description of the model and the physical stack of the simulated \ac{RRAM} can be found in the Supplementary Information, Section 2.1.

        In this work, current-pulse programming, a common approach in neuromorphic hardware \cite{GiacomoDPI2009, GiacomoACCCN2011}, is used to program the \ac{RRAM} device.
        Therefore, the analogue behaviour of the \ac{RRAM} is studied with current pulses of 10\,\textmu s time width and amplitudes from 50\,\textmu A to 300\,\textmu A with 10\,\textmu A steps. After each programming pulse, a read-out operation is performed in current mode. During this read-out operation, a current pulse with 10\,\textmu A amplitude and 10\,\textmu s pulse width is applied to the \ac{RRAM} and the voltage drop across the cell is measured to calculate the corresponding conductance.
        
        To qualitatively discuss the impact of \ac{RRAM} stochasticity on device programming, the filament-related parameters responsible for the conductance variation are varied by 1\,$\mathrm{\sigma}$ from their mean values, as shown in the Supplementary Information, Table 2.
        
        The quantised values of the \ac{RRAM} were obtained with current pulses of 10\,\textmu s time width and amplitudes as in Table~\ref{tab:current pulse values}. Note that these values were obtained with partial set operations starting from the \ac{HRS} and reaching different intermediate \ac{LRS}.

        \begin{table}[t]
            \caption{Current pulse amplitudes for 3-bits quantisation}\label{tab:current pulse values}
            \centering
            \begin{tabular}{*{9}{cc}}
                \toprule
                States & Current pulse amplitude \\
                \midrule
                $HRS$ & 10\,\textmu A\\
                $LRS_{1}$ & 50\,\textmu A\\
                $LRS_{2}$ & 80\,\textmu A\\
                $LRS_{3}$ & 110\,\textmu A\\
                $LRS_{4}$ & 140\,\textmu A\\
                $LRS_{5}$ & 180\,\textmu A\\
                $LRS_{6}$ & 230\,\textmu A\\
                $LRS_{7}$ & 300\,\textmu A\\
                \bottomrule
            \end{tabular}
        \end{table}

    \subsection{Quantisation-Aware training}
        \label{sec:methods:RRAM quantisation}
        The quantisation of the synapses throughout the training process has to be considered to adequately model the \ac{RRAM}-based synapses. Weight quantisation is incorporated into the training phase using the \ac{qat} process \cite{JMLR:v18:16-456}. Since the quantisation operation is non-differentiable, a surrogate quantisation operation is used to perform the weight quantisation. This means that quantisation is applied in the forward pass, but is replaced by an approximate differentiable function in the backward pass.
        \begin{equation}
            \label{eq:qat:round}
            w_q = \mathrm{sround}\left(\frac{w}{s_w}\right)
        \end{equation}
    
        \begin{equation}
            \label{eq:qat:scale}
            s_w = \frac{\mathrm{max}\left(|w|\right)}{2^{N-1}-1}
        \end{equation}
        
        The quantisation operator is defined in the equation \ref{eq:qat:round}, where the trained floating-point weight value $w$ is scaled by a factor $s_w$ defined in the equation \ref{eq:qat:scale}, where N is the number of bits to be considered. The scaled weight is then applied to a stochastic rounding operator \cite{croci2022stochastic}. Since the gradient of the rounding is approximately zero over the domain, the \ac{ste} surrogate gradient is applied to it \cite{bengio2013estimating,Zhang_2022_CVPR}.

    \section{Results}
    \label{sec:bruno}
    \subsection{Training algorithm}
        \label{sec:bruno:algorithm}
\begin{figure}[t]
            \centering
            \includegraphics[]{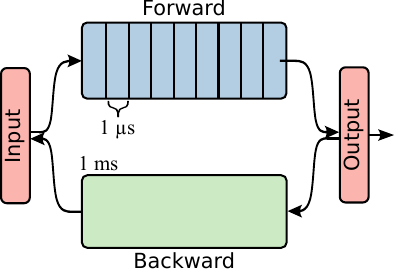}
            \caption{Schematic of the algorithm for computing the forward and backward passes with different time steps. During the forward pass (blue box) the simulation operates on 1\,\textmu s timescale. In the backward pass (green box) the gradient is calculated on 1\,ms timescale.}
            \label{fig:bruno_algorithm}
        \end{figure}

        Training physically accurate neuron models with different temporal dynamics can cause instabilities at simulation timescales of $\approx$1\,ms. In addition, training at finer timescales of around 1\,\textmu s can increase computational resources by three orders of magnitude.
        \ac{bruno} uses a dual timescale for the forward pass and the gradient computation. Figure \ref{fig:bruno_algorithm} illustrates the training procedure. In the forward pass, the network simulation runs with a fine-grained simulation time step of 1\,\textmu s. However, the gradient calculation is performed with a 1\,ms time step. This reduces the size of the unrolled computational graph during backpropagation by reducing the total number of time steps in the backward pass. To achieve the dual timescale training, the calculation of the internal state of the neuron can be computed on both time steps and use the higher timescale gradient as a surrogate gradient for the 1\,\textmu s time resolution value (Algorithm \ref{alg:bruno}).

        To validate \ac{bruno}, the \ac{felif} model described in section \ref{sec:methods:felif} is used. The internal state of the neuron $s^t = \{V^t, P^t\}$ is defined as the voltage and polarization over time. The algorithm \ref{alg:bruno} shows the pseudocode of the training method. In the forward pass, the value of the neuron state at time t ($s^t$) is calculated with microsecond ($s_{\mu s}$) and millisecond ($s_{ms}$) time resolution. By applying the function $s_{ms} + detach(s_{\mu s} - s_{ms})$, the state of the neuron is equal to $s_{\mu s}$. However, as the subtraction $s_{\mu s} - s_{ms}$ is detached from the computational graph, the gradient flows throughout the millisecond time scale state ($s_{ms}$). The final voltage $v_{mem}$ obtained from $s_{ms}$ is then applied to a threshold operator $sg$ that applies a surrogate gradient \cite{10342587}. When the neuron fires, the state is reset. The software implementation of \ac{bruno} is publicly available on Github \footnote{\url{https://github.com/bics-rug/ELEANOR}}
        
        \begin{algorithm}
            \caption{\ac{bruno}}\label{alg:bruno}
            \begin{algorithmic}
                \For{$t = 1, \dots, T$}
                    \State $s_{\mu s}^0 \gets s^{t-1}$
                    \For{$i = 1, \dots, 1000$}
                        \State $s_{\mu s}^i \gets$ compute $s$ based on equations \ref{eq:v_derivative_circuit} - \ref{eq:tau} with 1\,\textmu s $\Delta$t
                    \EndFor
                    \State $s_{ms} \gets$ compute $s$ based on equations \ref{eq:v_derivative_circuit} - \ref{eq:tau} with 1\,ms $\Delta$t
                    \State $s^t \gets s_{ms} + detach(s_{\mu s} - s_{ms})$
        
                    \State spike $\gets sg(v_{mem}^t - v_{thr})$
                    \State reset $s^t$ in case of spike
                \EndFor
            \end{algorithmic}
        \end{algorithm}
    
    \subsection{Verification and benchmarking}
    \label{sec:res}

    To verify the functionality of \ac{bruno} on a physically accurate neuron model for future hardware deployment, first, the behaviour of the \ac{felif} Python model used during training is verified to match the \ac{spice} simulations of the \ac{felif} circuit. Second, the \ac{RRAM} stochasticity and quantisation for the synapse implementation are analysed. Finally, a comparison and analysis of \ac{bruno} and \ac{bptt} is performed, examining the differences in training time and memory consumption. This is followed by benchmarking an \ac{snn} on different spatio-temporal datasets, with the results compared across different cases involving different quantisation levels for the weights and different neuron models.

    \subsubsection{Hardware model and verification}
    \label{sec:res:model_verification}
\begin{figure}[t]
            \centering
            \begin{subfigure}[b]{0.49\textwidth}
                \centering
                \includegraphics[width=\textwidth]{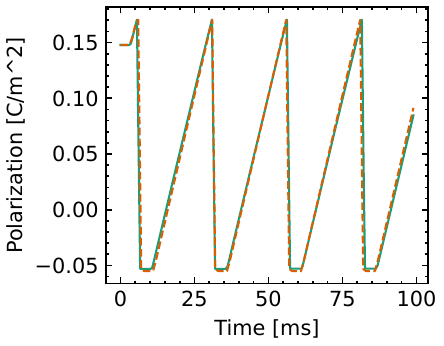}
                \caption{}\label{fig:SPICE_python:a}
            \end{subfigure}
            \begin{subfigure}[b]{0.49\textwidth}
                \centering
                \includegraphics[width=\textwidth]{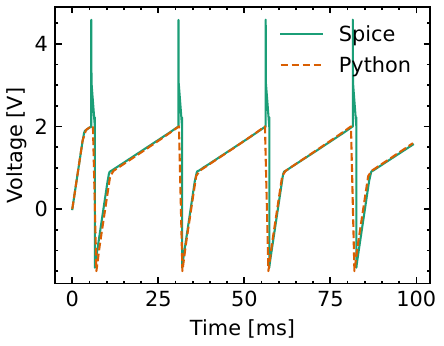}
                \caption{}\label{fig:SPICE_python:b}
            \end{subfigure}
            \caption{Comparison of the membrane potential transient between the \ac{spice} simulation of the neuron circuit and the model used in the network simulations with a DC input current of 308\,pA. Except for the spike amplitude, the polarization (a) and the voltage (b) match between the Python and \ac{spice} simulations. The amplitude of the spike in the network is negligible as it is registered as an event in any case.}
            \label{fig:SPICE_python}
        \end{figure}

    A comparison of the \ac{spice} circuit simulations and the behavioural model (equations \ref{eq:v_derivative_circuit}-\ref{eq:tau}) shows a close match in the transient of the ferroelectric polarization (Figure \ref{fig:SPICE_python:a}) as well as in the membrane potential of the neuron (Figure \ref{fig:SPICE_python:b}). While the network simulation resets the membrane potential as soon as the firing threshold is crossed, the circuit needs to reset the physical device, resulting in an overshoot in the voltage. However, since it happens during the refractory period, it does not affect the operation of the neuron in the network simulations.

    \subsubsection{Synaptic Stochastic and Quantisation analysis}
        \label{sec:results:stochasticRRAM}
    
        Due to their non-volatility and small footprint, \acp{RRAM} were chosen as the artificial synapses for the neural networks used in the \ac{bruno} benchmarking. They operate on the principle of the formation or rupture of conductive filaments, resulting in the switching between \ac{LRS} and \ac{HRS}. However, due to the physical processes leading to this switching (the random formation and rupture of the conductive filaments during the SET and RESET processes \cite{ielmini2016resistive}), \ac{RRAM} devices exhibit inherent stochasticity, leading to \ac{C2C} variability. As a result, adjacent conductance states may overlap, making them indistinguishable. Therefore, quantisation is required to define a limited number of discrete, non-overlapping conductance levels to ensure distinguishable weights/conductance states. The inherent stochastic behaviour of \ac{RRAM} devices was incorporated into the device simulations using stochastic rounding as described in Section \ref{sec:methods:RRAM quantisation}.
        
\begin{figure}[t]
            \centering
            \begin{subfigure}[b]{0.49\textwidth}
                \centering
                \includegraphics[width=\textwidth]{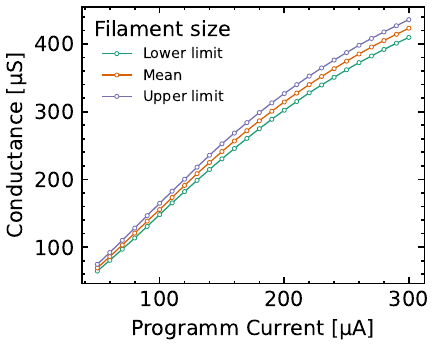}
                \caption{}
                \label{fig:AnalogWeightUpdate}
            \end{subfigure}
            \begin{subfigure}[b]{0.49\textwidth}
                \centering
                \includegraphics[width=\textwidth]{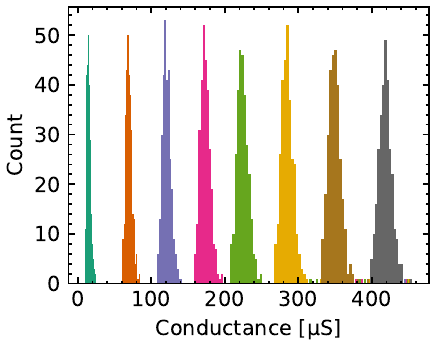}
                \caption{}
                \label{fig:3bits quantisation}
            \end{subfigure}
            \caption{\ac{RRAM} state programming and device conductance. (a) Device conductance based on the programming current applied, for a mean filament size of 0.4\,nm length and 45\,nm filament radius, an upper limit of 0.36\,nm length and 49\,nm filament radius, and a lower limit of 0.44\,nm length and 41\,nm filament radius. (b) Distribution of the different 3-bit quantised states.}
            \label{fig:RRAM graphs}
        \end{figure}

        Before quantizing the \ac{RRAM} programmed states, it is essential to investigate the \ac{RRAM} analogue weight update behaviour and assess its linearity.
        Figure \ref{fig:AnalogWeightUpdate} illustrates the change in conductance in response to an increase in the amplitude of the programmed current pulse, as well as the maximum variability of the programmed conductance state due to the stochastic switching behaviour \cite{nardi2012resistive, ambrogio2014statistical}. It can be observed that a quasi-linear change in conductance follows the application of each programming pulse, with an increase in the amplitude of the current pulse. However, the \ac{C2C} variability limits the number of states that can be reliably read, since the same analogue conductance state can be obtained with different current values. As a result, the distributions of different states overlap, preventing a correct reading. To reliably extract distinguishable \ac{RRAM} programmed states from the quasi-analogue weight update, state quantisation is performed. Based on the data in Figure \ref{fig:AnalogWeightUpdate}, 3-bit states have been quantised. Figure \ref{fig:3bits quantisation} illustrates the distribution of 3-bit quantised states obtained from \ac{MC} simulations, which included 7 \ac{LRS} states extracted from the analogue weight update curve and 1 initial \ac{HRS} state, as explained in Section \ref{sec:methods:RRAM}. The distribution of conductance within 2\,$\mathrm{\sigma}$ over the selected programming pulses shows a clear distribution across the entire conductance range. The distribution of adjacent conductance states shows a distinct gap with an average of 20\,\textmu S, which is sufficient to be detected by standard \ac{cmos} circuitry.        
            
        \subsubsection{Benchmark}
        \label{sec:res:benchmark}
        To evaluate the performance of \ac{bruno} on a high-temporal-resolution model such as the \ac{felif} neuron, two spatio-temporal datasets were used. Firstly, the \ac{jsb} music chorales dataset \cite{boulanger2012modeling} was used for music generation, with notes represented as input spikes at each time step. Secondly, the Neuromorphic Braille Reading Dataset \cite{muller2022braille}. This consists of 5,400 Braille reading samples recorded by an Omega 3 robot with a sensorised fingertip that moves over 3D-printed Braille letters \cite{muller2022braille}. Furthermore a comparison between training with \ac{bptt} and \ac{bpttc} or \ac{bruno} were made on both datasets.

\begin{figure}[htpb]
            \centering
                \centering
                \includegraphics[width=\textwidth]{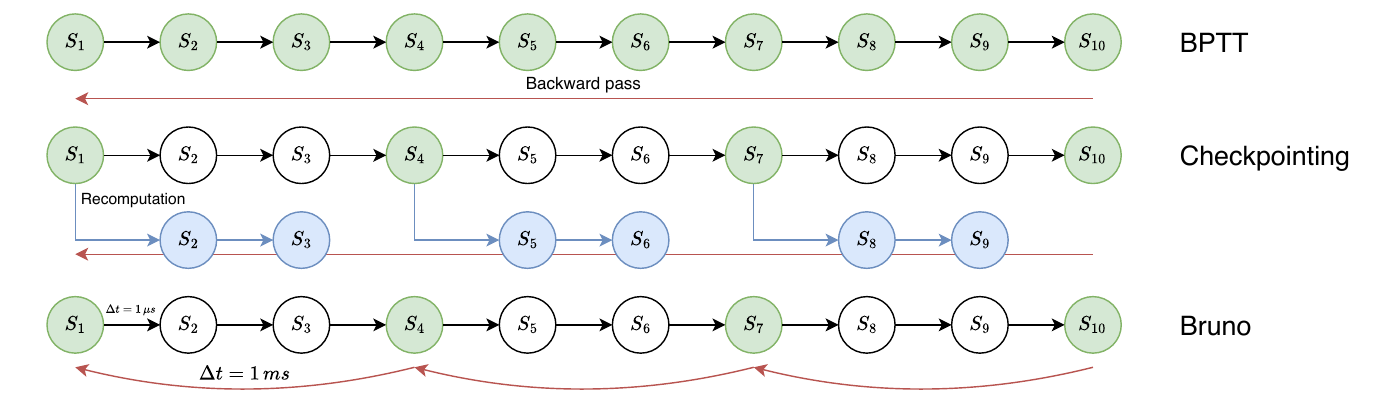}
                \caption{Comparison of forward and gradient calculations of \ac{bptt}, checkpointing and \ac{bruno}. The green nodes represent points saved in memory and the blue nodes represent those that has be recomputed during the backward pass with checkpointing. In \ac{bptt} the internal states of the model are stored at each time step, which increases memory consumption. Checkpointing provides a trade-off between time and memory, by storing key state points; however, it must recompute the intermediate states during backpropagation, which increases computation time. In the case of \ac{bruno}, internal states at at key times are stored for the backward pass as with checkpointing. Furthermore, gradient computation does not recompute all intermediate states, as gradient computation occurs at a different timescale. This reduces the training time and memory consumption compared to \ac{bptt} and checkpoints.}
                \label{fig:bruno_diagram}
        \end{figure}

        \paragraph{To \ac{bruno} or not to \ac{bruno}}
        Training a neural network for long sequences ($>$1M) with \ac{bptt} can cause memory and time problems during training, as it has to store internal states for every time step. Methods like checkpointing \cite{bencheikh2024optimal} can reduce the memory consumption by storing only key states, but during the backpropagation the intermediate states that were not stored need to be recomputed. In the case of \ac{bruno}, there is no need to recompute the intermediate states, since it backpropagate with a higher time step. (Fig. \ref{fig:bruno_diagram}).

        \begin{figure}[ht]
            \centering
                \centering
                \includegraphics{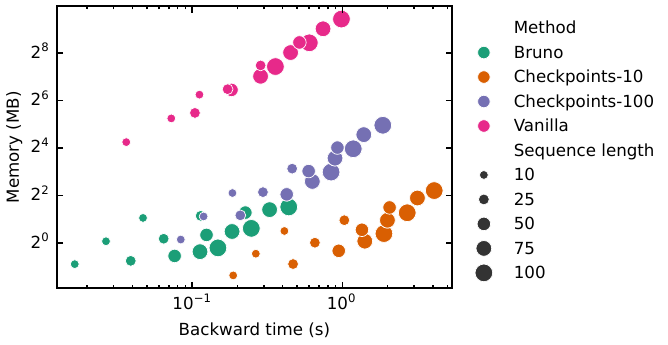}
                \caption{Memory and Time analysis between different algorithms. Each method is represented with a different color, where the size of the dots are the sequence length of input sequence. Bruno has an improvement in performance compare in the memory saving and gradient calculation time compare to method like checkpoint, that provides a tradeoff between time and memory resources.}
                \label{fig:memory_analysis}
        \end{figure}

        Figure \ref{fig:memory_analysis} compares the training time and memory consumption of training a neural network with difference size (128, 256 and 512 neurons) with different sequence lengths for \ac{bptt}, \ac{bpttc} and \ac{bruno}. In the case of \ac{bpttc} different number of checkpoints where used by trading memory and training time (since it need to recompute more nodes). \ac{bruno} improves both training time and memory utilization compared to the other methods.  

\begin{figure}[ht]
            \begin{subfigure}[t]{0.49\textwidth}
                \centering
                \includegraphics[width=\textwidth]{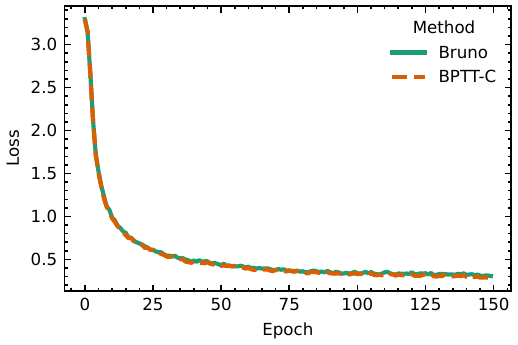}
                \caption{}
                \label{fig:loss_checkpoint}
            \end{subfigure}%
            ~
            \begin{subfigure}[t]{0.49\textwidth}
                \centering
                \includegraphics[width=\textwidth]{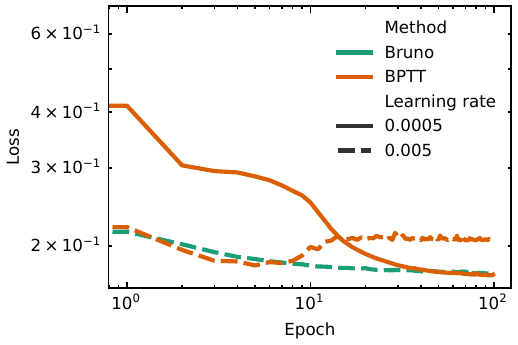}
                \caption{}
                \label{fig:loss_landscape}
            \end{subfigure}%
            \caption{Comparison between training with \ac{bruno} or \ac{bptt}/C on the \ac{jsb} and Braille reading datasets. Figures a) and b) shows the loss evolution on both datasets for each method. (a) Loss between Bruno and \ac{bpttc} on the Braille reading dataset. (b) Loss between Bruno and \ac{bptt} on the \ac{jsb} dataset. In the case of \ac{bptt} multiple learning rate where used, without an improve in performance in comparison with \ac{bruno}. All the parameters where the same for each method.}
        \end{figure}

        Although the gradient calculation with \ac{bruno} can differ from \ac{bptt} and \ac{bpttc}, we investigate its convergence and compare it with \ac{bptt} and \ac{bpttc} by training a network with the \ac{jsb} chorales and Braille reading datasets. The results in the braille recognition task show a similar loss evolution with respect to \ac{bptt} without a loss in performance (Fig. \ref{fig:loss_checkpoint}). In the case of the \ac{jsb} dataset \ac{bruno} outperform training with \ac{bptt} with the same set of parameters.

        \paragraph{Bach Chorales Music prediction}
        The \ac{jsb} dataset \cite{boulanger2012modeling,bohnstingl2022online} is a music prediction task, where the objective is to predict the next musical note based on previous inputs. A single layer network of \ac{felif} neurons were used, followed by a linear layer with sigmoid activation. The output of the network corresponds to the probability of each note to be play at each time step. 
        The loss function used is the sigmoid cross-entropy defined at each time step.

        \begin{equation}
             L = - \sum_{c=1}^Mylog(p)
        \end{equation}
        Where $M$ is the number of notes, $y\in\{0,1\}$ represents if the note is active, and the $p$ is the probability that each note will be activated in each time step provided by the sigmoid activation function.
        
\begin{figure}
            \begin{subfigure}[t]{0.49\textwidth}
                \centering
                \includegraphics{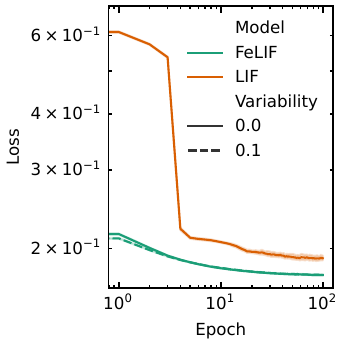}
                \caption{}
                \label{fig:jsb-loss}
            \end{subfigure}
            \begin{subfigure}[t]{0.49\textwidth}
                \centering
                \includegraphics{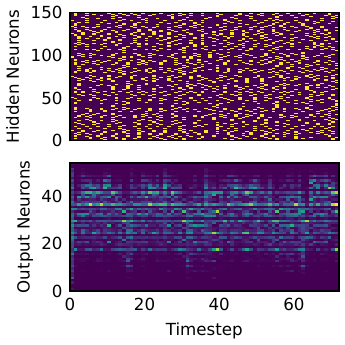}
                \caption{}
                \label{fig:jsb-output}
            \end{subfigure}
            \caption{Comparison between training a network composed of \ac{felif} and \ac{lif} neurons on the \ac{jsb} music prediction task. (a) Loss over 100 epochs for the \ac{jsb} dataset with the \ac{felif} neuron model using Bruno and \ac{lif}. In the case of the \ac{felif} neuron, a 10\% of variability on the \ac{fecap} parameters were added (Supplementary Information, Section 1.1). (b) Output spikes and probability of the network with \ac{felif} neurons. The upper figure represents the firing of each neuron in the hidden layer at each time step, the figure bellow has the output probability for each note at each time step.}
            \label{fig:jsb-train}
        \end{figure}

        The \ac{felif} neuron trained with \ac{bruno} shows good training capabilities on this datasets, surpassing a network with \ac{lif} neurons (Table \ref{tab:loss_jsb} and Figure \ref{fig:jsb-train}). For the network with \ac{lif} neurons, a decay of 0.4 \cite{osttp} and learning rate of 0.05 were used. For the \ac{felif} neurons, a $I\_dsc=10\text{\,pA}$, learning rate of 0.005 and threshold 2.0\,V where used.

        \begin{table}[tbh]
            \caption{Binary crosentropy loss for different neuron models on the JSB dataset for music prediction, where the network with \ac{felif} neurons are trained with \ac{bruno}.}\label{tab:loss_jsb}
            \centering
            \begin{tabular}{lcc}
            \toprule
            Network & $\mu$  & $\sigma$ \\
            \midrule
            LIF & 0.189365 & $3.795\times 10^{-3}$ \\
            FeLIF & 0.174426 & $5.54\times10^{-4}$ \\
            FeLIF (10\% variability) & 0.174613 & $3.37\times10^{-4}$ \\
            \bottomrule
            \end{tabular}
        \end{table}

        \paragraph{Braille letter recognition}
        \label{sec:benchmark:braille}
        \begin{figure}[t]
            \centering
            \begin{subfigure}[b]{0.25\textwidth}
                \centering
                \includegraphics[]{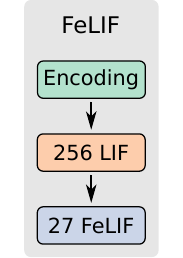}
                \caption{}
                \label{fig:networks_all:a}
            \end{subfigure}
            \begin{subfigure}[b]{0.25\textwidth}
                \centering
                \includegraphics[]{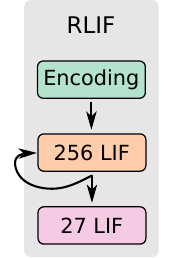}
                \caption{}
                \label{fig:networks_all:b}
            \end{subfigure}
            \begin{subfigure}[b]{0.25\textwidth}
                \centering
                \includegraphics[]{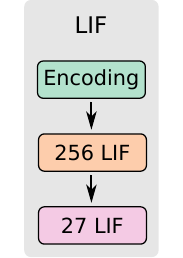}
                \caption{}
                \label{fig:networks_all:c}
            \end{subfigure}
            
            \caption{Network architectures used in the benchmarks. All networks contain an encoding layer, a hidden layer with 256 \ac{lif} neurons, and an output classification layer. (a) Feedforward network with \ac{felif} neuron in the output layer. (b) Feedforward network with the \ac{lif} neurons in the output layer. (c) Network where the \ac{lif} hidden layer has an explicit recurrent connection and the output neurons are \ac{lif} neurons.}
            \label{fig:networks_all}
        \end{figure}
    
        To evaluate the performance of the \ac{felif} neuron and \ac{bruno} on this dataset, three different networks were constructed. The three networks have an input encoding layer that expands the input dimension to 256, a hidden layer of 256 \ac{lif} neurons, which transforms the expanded analogue sensor values of the Braille letters into spikes. The output layer consists on 27 neurons, representing each of the Braille letters, including the space. The first network uses \ac{felif} neurons in the output layers (Figure \ref{fig:networks_all:a}), while the second uses \ac{lif} neurons (Figure \ref{fig:networks_all:b}). Moreover, for spatio-temporal data, \ac{lif} neuron networks may have difficulty learning long temporal dependencies due to the short-term dynamics \cite{9731734}. Adding explicit recurrence can improve the performance on spatio-temporal data \cite{quintana2024etlp}. For this reason, the third network incorporates recurrent connection weights in the hidden layer (Figure \ref{fig:networks_all:c}). A \ac{hpo} was performed to train the three network architectures \ac{snn} with different synaptic quantisation values. The final parameters used can be found in Table \ref{tab:optuna_parameters}.
    
\begin{figure}[htb]
            \centering
            \includegraphics[width=\textwidth]{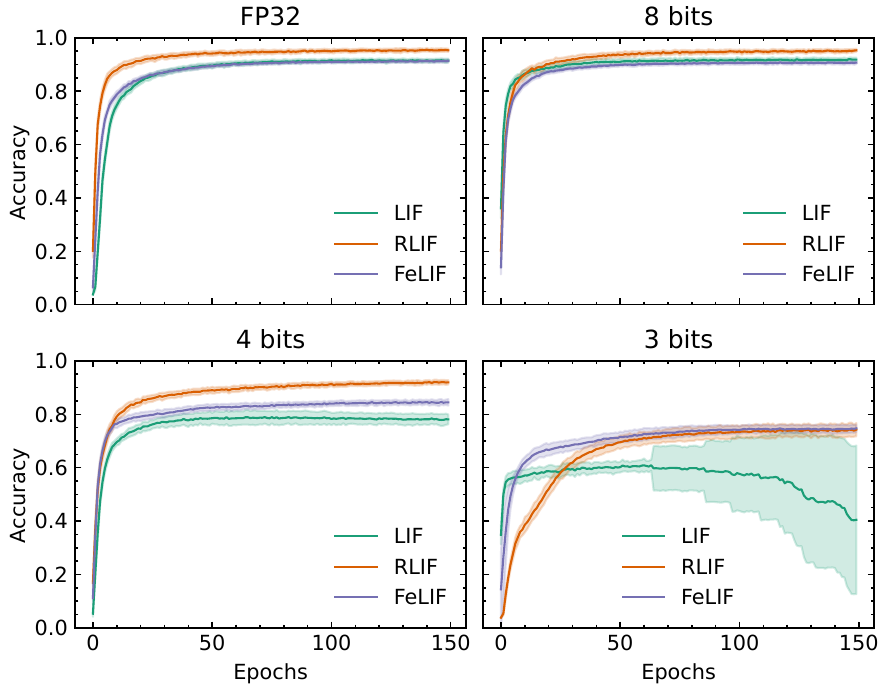}
            \caption{Accuracy on the Braille dataset with different quantisation levels and networks. Each plot has different quantisation levels (Floating-point, 8, 4, and 3 bits). In each plot, the three different network architectures are shown, trained with different initialisation seeds using the hyperparameters obtained using \ac{hpo}}
            \label{fig:results}
        \end{figure}

        \paragraph{Quantisation and model comparison}
        The next validation performed was the comparison between different synaptic quantisation levels and neuron models. As explained earlier, three different networks were created comparing the \ac{felif} neuron model with a feedforward and recurrent \ac{lif}. For each network and quantisation value, \ac{hpo} was performed using the Optuna library \cite{optuna_2019}. Once the optimal hyperparameters were obtained (Table \ref{tab:optuna_parameters}) for each experiment (network and quantisation), 50 simulations were performed with different initialisation values. Figure \ref{fig:results} and Table \ref{tab:accuracy} show the final accuracy obtained. Without quantisation, the feedforward \ac{felif} network achieves a similar performance as the feedforward \ac{lif} network with an accuracy of $91.44\%\pm0.73$ compared to $91.34\%\pm0.59$ for the \ac{felif}. In contrast, the recurrent network has an accuracy of $95.30\%\pm0.61$. However, as the quantisation level of the network increases, the performance of the feedforward \ac{felif} becomes better than that of the \ac{lif} with recurrence, obtaining $74.73\%\pm1.37$ and $74.26\%\pm2.48$ respectively, compared to the feedforward \ac{lif} which drops drastically in performance to $40.48\%\pm27.84$.
    
        \begin{table}[tbh]
        \caption{Parameters used for network simulations with different quantisation levels and models. The $alpha_{hid/out}$ parameter is the membrane decay constant for the hidden and output layers. The $\beta_{hid/out}$ parameters are the synapse decay constants for the hidden and output layers. The \ac{felif} V\textsubscript{thr} parameter is the firing threshold of the \ac{felif} neuron.}\label{tab:optuna_parameters}
            \centering
            \begin{threeparttable}
            \begin{tabular}{llllllll}
                \toprule
                Quant. & Model & $\alpha_{hid}$ & $\beta_{hid}$ & $\alpha_{out}$ & $\beta_{out}$ & LR\tnote{1} & FeLIF V\textsubscript{thr} \\
                \midrule
                \multirow{ 3}{*}{3} & LIF & 0.662 & 0.703 & 0.565 & 0.696 & $3.429 \times 10^{-2}$ & - \\
                 & RLIF & 0.603 & 0.310 & 0.354 & 0.295 & $1.478 \times 10^{-3}$ & - \\
                 & FeLIF & 0.468 & 0.735 & - & - & $7.840 \times 10^{-4}$ & 3.039 \\
                \multirow{ 3}{*}{4} & LIF & 0.226 & 0.245 & 0.865 & 0.834 & $3.276 \times 10^{-3}$ & - \\
                 & RLIF & 0.599 & 0.710 & 0.390 & 0.349 & $6.055 \times 10^{-3}$ & - \\
                 & FeLIF & 0.456 & 0.322 & - & - & $3.716 \times 10^{-3}$ & 2.544 \\
                \multirow{ 3}{*}{8} & LIF & 0.230 & 0.302 & 0.901 & 0.591 & $1.283 \times 10^{-2}$ & - \\
                 & RLIF & 0.186 & 0.537 & 0.347 & 0.846 & $6.766 \times 10^{-3}$ & - \\
                 & FeLIF & 0.682 & 0.662 & - & - & $4.387 \times 10^{-3}$ & 2.928 \\
                \multirow{ 3}{*}{FP} & LIF & 0.959 & 0.202 & 0.716 & 0.600 & $3.006 \times 10^{-3}$ & - \\
                 & RLIF & 0.343 & 0.406 & 0.764 & 0.874 & $6.174 \times 10^{-3}$ & - \\
                 & FeLIF & 0.299 & 0.147 & - & - & $2.628 \times 10^{-3}$ & 3.388 \\
                \bottomrule
            \end{tabular}
            \begin{tablenotes}
            \item[1] Learning rate
            \end{tablenotes}
            \end{threeparttable}
        \end{table}
    
        \begin{table}[htb]
            \caption{Test accuracy obtained for different quantisation levels and network architectures on the neuromorphic Braille letter reading dataset.}\label{tab:accuracy}
            \centering
            \begin{tabular}{ccccccccc}
            \toprule
            \multirow{2}{*}{Network} & \multicolumn{2}{c}{FP32} & \multicolumn{2}{c}{8 bits} & \multicolumn{2}{c}{4 bits} & \multicolumn{2}{c}{3 bits} \\
            \cmidrule{2-9}
             & $\mu$  & $\sigma$ & $\mu$  & $\sigma$ & $\mu$ & $\sigma$ & $\mu$ & $\sigma$ \\
            \midrule
            LIF & 91.44 & 0.73 & 91.81 & 0.73 & 78.10 & 1.85 & 40.48 & 27.84\\
            RLIF & 95.30 & 0.61 & 95.29 & 0.59 & 91.96 & 0.84 & 74.26 & 2.48 \\
            FeLIF & 91.34 & 0.59 & 90.69 & 0.57 & 84.49 & 1.22 & 74.73 & 1.37 \\
            \bottomrule
            \end{tabular}
        \end{table}
    
\section{Discussion and conclusion}
    This work presents a method for training non-linear, high-temporal neuron models derived from the physical mechanisms and behaviour of analogue hardware. Due to the physics-based nature, the simulation of these models can exhibit instability when matching application time steps of $\approx$1\,ms. The proposed training method is comparable to standard techniques, such as \ac{bptt}, in the training of the \ac{felif} model, thereby reducing the training time and memory consumption. The \ac{felif} neuron model was used to validate the training algorithm. The model was trained with and without \ac{bruno} on the \ac{jsb} dataset for music prediction and the Braille reading dataset, both datasets used for benchmarking spatio-temporal patterns. The results show that training with \ac{bruno} improves training time, memory requirements, and accuracy. Achieving a reduction between 97-99\% of the peak memory usage and 50-60\% for reduction in training time, compared to \ac{bptt}. With an accuracy of $91.34\%\pm0.59$ on the braille dataset. By contrast, training with \ac{bpttc} involves a trade-off in terms of computational cost, with increased training time and reduced memory usage.
    
    Benchmarking with feedforward and recurrent \ac{lif} networks was performed in addition to the validation of the training procedure with \ac{bptt} and \ac{bpttc} methods. This benchmark was also performed with synaptic quantisation for its possible future implementation on hardware with \ac{RRAM} synapses. Different quantisation levels were performed, for 8, 4, and 3 bits, where \acf{hpo} was performed to obtain the optimal set of hyperparameters for each network and quantisation level. As a result of this benchmark, we show that a feedforward \ac{felif} network has comparable results to a feedforward \ac{lif} network with $91.34\%\pm0.59$ and $91.44\%\pm0.73$ respectively. However, for the quantisation network, the \ac{felif} outperforms the feedforward \ac{lif} network and has similar performance with a recurrent \ac{lif} network with 3-bit quantisation with $74.73\%\pm1.37$ and $74.23\%\pm2.48$ respectively. In contrast, the mean accuracy obtained for the feedforward network is $40.48\%\pm27.84$, which could be due to a high learning rate for the quantised weights, since the network started to lose performance drastically on the last epochs.
    
    Thanks to the reduction in resources need for training compared to methods like \ac{bptt} or \ac{bpttc}, \ac{bruno} opens up the possibility of simulating and training neural models with complex dynamics that can be used for spatio-temporal pattern recognition. These neural models have the potential to be integrated into hardware with synaptic quantisation, which could provide performance comparable to that of a recurrent \ac{lif} network, while not requiring additional recurrent connections, and thus fewer trainable parameters.

\section*{Acknowledgements}
We thank Pedro F. Ortiz Costa and Dr. Francesco Malandrino, Jun. Prof. David Kappel for their support and fruitful discussions. We thank the Center for Information Technology of the University of Groningen for their support and for providing access to the Hábrók high performance computing cluster. This work was partially supported by the European Union and the European Research Council (ERC) through the European’s Union Horizon Europe Research and Innovation Programme under Grant Agreements No 101042585 and 101207193. Views and opinions expressed are however those of the authors only and do not necessarily reflect those of the European Union, the European Research Council or the European Research Executive Agency. Neither the European Union nor the granting authority can be held responsible for them. The authors would like to acknowledge the financial support of the CogniGron research center and the Ubbo Emmius Funds (Univ. of Groningen).

\bibliographystyle{ieeetr} 
\bibliography{references}



\section*{Supplementary material}

# Supplemental Material for: "Bruno: Backpropagation Running Undersampled for Novel device Optimization"

May 23, 2025

## 1 Resistive switching non-volatile devices (RRAM)

### 1.1 Resistive switching mechanism

Memristive devices based on resistive switching RRAM, due to their capacity to support non-volatile storage of synaptic weight, have been selected for use as artificial synapses in Backpropagation Running Undersampled for Novel device Optimization (BRUNO). In particular, this work employs the valence change mechanism (VCM) as the primary switching mechanism for RRAM devices. The formation or dissolution of the conductive filament can be achieved by moving the oxygen vacancies, which are usually generated by oxygen exchange at the electrode during the forming process, back and forth in the region between the filament tip and bottom electrode (BE) (or "disc" region) by switching the electric field between the top electrode (TE) and BE as shown in Figure 1. These operations are designed as SET and RESET, respectively, and consequently, the RRAM cell is switched between low resistive states (LRS) and high resistive state (HRS). The simulation of the switching behaviors of RRAM devices is conducted using the Jülich Aachen Resistive Switching Tool Jülich Aachen Resistive Switching Tool (JART) VCM physical compact model [1]. In the JART VCM model, the RRAM cell is based on a $HfO_x/TiO_x$ material system. As a set of SPICE-level physical models, this model allows us to design functional complementary metal-oxide semiconductor (CMOS)-based circuits and combine them with the RRAM model to construct our testbench in cadence.

### 1.2 Simulation of state quantization

State quantization is necessary to extract the distinguishable weights, that is, the maximum number of non-overlapping states that can be confidently programmed in the proposed RRAM device when accounting for Cycle to Cycle (C2C) variability. Regarding the quantization of the RRAM, programming currents were selected to program multiple LRS in addition to the initial HRS.

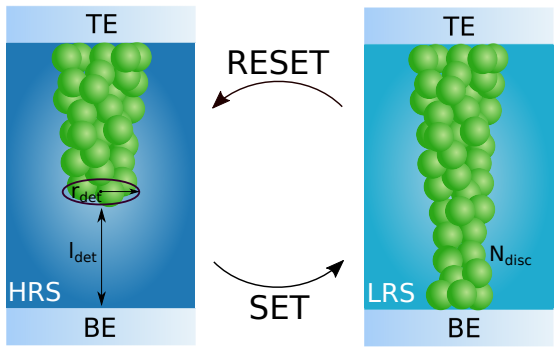

Figure 1: RRAM switching process in disc region in form of formation and dissolution of the filament caused by forward and backward moving of oxygen vacancy (represented by the green spheres) under electric field between TE and BE.

| Parameter | Mean value | $\sigma$ |
|---|---|---|
| $N_{disc,max}$ | $20 \times 10^{26}/m^3$ | $2 \times 10^{26}/m^3$ |
| $N_{disc,min}$ | $0.008 \times 10^{26}/m^3$ | $0.004 \times 10^{26}/m^3$ |
| $r_{det}$ | 45 nm | 4 nm |
| $l_{det}$ | 0.4 nm | 0.04 nm |

Table 1: Variability considered for filament-related parameters

Monte Carlo Monte Carlo (MC) simulations are conducted for each state, with 300 samples, to assess the impact of varying physical parameters, including maximum oxygen vacancy concentration in the disc ($N_{disc,max}$), minimum oxygen vacancy concentration in the disc ($N_{disc,min}$), length of the disc ($l_{det}$) and radius of filament ($r_{det}$). The values of $N_{disc,max}$, $N_{disc,min}$, $l_{det}$, $r_{det}$ in each MC sample are selected from their respective Gaussian distributions to ensure that the quantities remain within reasonable ranges. The utilized values are listed in Table 1.

As shown in Fig. 2, for each SET pulse amplitude, there is still a limited number of cycles that exhibit higher programmed conductance values, forming a tail of the distribution ($3\sigma$). The reason for this distribution tail could be related to the simulation setup. Since all physical filament related parameters are randomly selected from a range of value distribution (here is Gaussian distribution), there is highly chance that the combination of selected $N_{disc,max}$, $l_{det}$, $r_{det}$ leads to a higher oxygen vacancy concentration in disc region during SET and thus leads to these higher conductance states. Additionally, narrower the initial HRS state distribution can be observed in Fig. 2. The initial filament physical parameter set up and small current pulse to the RRAM cell attribute to this narrower distribution. First of all, initial value of $l_{det}$, $r_{det}$ converge in a narrower Gaussian distribution. Secondly, the current pulse amplitude applied

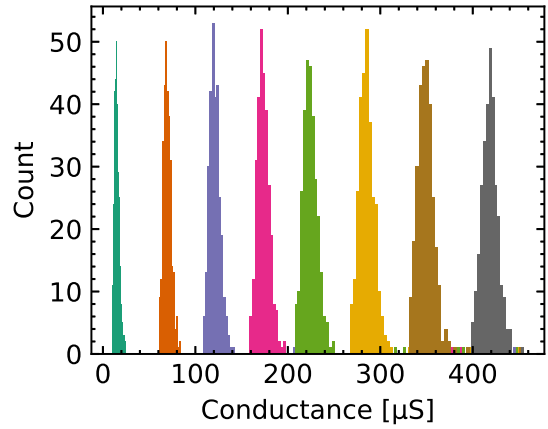

Figure 2: RRAM quantization figure.

to TE in this case is 10 µA, which is too small to trigger SET event. As a result, the extracted conductance value is close to the initial HRS.



\end{document}